\documentclass{sn-jnl}
\usepackage{graphicx}%
\usepackage{multirow}%
\usepackage{amsmath,amssymb,amsfonts}%
\usepackage{amsthm}%
\usepackage{mathrsfs}%
\usepackage[title]{appendix}%
\usepackage{xcolor}%
\usepackage{booktabs}
\usepackage{textcomp}%
\usepackage{manyfoot}%
\usepackage{tabularx}
\usepackage{booktabs}%
\usepackage{algorithm}%
\usepackage{algorithmicx}%
\usepackage{algpseudocode}%
\usepackage{listings}%
\usepackage[square,numbers]{natbib}
\theoremstyle{thmstyleone}%
\theoremstyle{thmstyletwo}%

\theoremstyle{thmstylethree}%

\begin{document}

\title[Deep Learning-Based Object Detection for Autonomous Vehicles: A Comparative Study of One-Stage and Two-Stage Detectors on Basic Traffic Objects]{Deep Learning-Based Object Detection for Autonomous Vehicles: A Comparative Study of One-Stage and Two-Stage Detectors on Basic Traffic Objects}


\author*[1]{\fnm{Bsher} \sur{Karbouj}}\email{karbouj@tu-berlin.de}

\author[1]{\fnm{Adam Michael } \sur{Altenbuchner}}

%\author[1]{\fnm{Erik} \sur{Sörqvist}}\email{soerqvist@tu-berlin.de}
\author[1]{\fnm{Jörg} \sur{Krüger}}
\affil*[1]{\orgdiv{Department of Industrial Automation Technology}, \orgname{Technical University of Berlin}, \orgaddress{\street{Pascalstr. 8-9}, \city{Berlin}, \postcode{10555}, \country{Germany}}}

%\affil[2]{\orgdiv{Department of Automation Technology}, \orgname{Frauhofer Institute IPK}, \city{Berlin}, \postcode{10587},\country{Germany}}

%\affil[3]{\orgdiv{Department}, \orgname{Organization}, \orgaddress{\street{Street}, \city{City}, \postcode{610101}, \state{State}, \country{Country}}}


\abstract{Object detection is a crucial component in autonomous vehicle systems. It enables the vehicle to perceive and understand its environment by identifying and locating various objects around it. By utilizing advanced imaging and deep learning techniques, autonomous vehicle systems can rapidly and accurately identify objects based on their features. Different deep learning methods vary in their ability to accurately detect and classify objects in autonomous vehicle systems. Selecting the appropriate method significantly impacts system performance, robustness, and efficiency in real-world driving scenarios. While several generic deep learning architectures like YOLO, SSD, and Faster R-CNN have been proposed, guidance on their suitability for specific autonomous driving applications is often limited. The choice of method affects detection accuracy, processing speed, environmental robustness, sensor integration, scalability, and edge case handling.
This study provides a comprehensive experimental analysis comparing two prominent object detection models: YOLOv5 (a one-stage detector) and Faster R-CNN (a two-stage detector). Their performance is evaluated on a diverse dataset combining real and synthetic images, considering various metrics including mean Average Precision (mAP), recall, and inference speed. The findings reveal that YOLOv5 demonstrates superior performance in terms of mAP, recall, and training efficiency, particularly as dataset size and image resolution increase. However, Faster R-CNN shows advantages in detecting small, distant objects and performs well in challenging lighting conditions. The models' behavior is also analyzed under different confidence thresholds and in various real-world scenarios, providing insights into their applicability for autonomous driving systems.}




\maketitle

\section{Introduction}\label{sec1}
The realization of autonomous driving is among the greatest endeavors of modern mobility. A key motivation behind autonomous driving is the desire to enhance the safety of all road users by preventing accidents caused by human error. Studies indicate that human error is a leading cause of traffic accidents, and autonomous vehicles can significantly mitigate this risk by eliminating distractions and ensuring precise control \cite{Channamallu_impact2023, Barabas_Current2017, cunningham2015autonomous}. Traffic flow can also be stabilized by avoiding unnecessary braking and acceleration, leading to not only a more comfortable journey but also to reduced $CO_2$ emissions \cite{jiang_Stop-and-Go2021}. Additionally, mobility is improved for individuals who rely on assistance, such as the elderly and people with disabilities, thereby alleviating the burden on caregivers and increasing their independence. This enhanced mobility allows these individuals to participate more fully in social and economic activities, improving their overall quality of life. \newline
Autonomous driving ideally relies on incoming multimodal input data from the environment, with image data being among the most crucial sources. To extract the necessary information from image data, object detection is employed, which is a subfield of machine vision. Object detection in autonomous driving presents unique challenges that distinguish it from general computer vision tasks, including real-time processing requirements, diverse environmental conditions, the long-tail problem, multi-scale detection, occlusion and overlap issues, and its safety-critical nature.  To address these challenges, various deep learning-based object detection models have been proposed and adapted for autonomous driving applications. \newline
Despite significant progress in recent years, object detection remains a challenge, particularly in unclear traffic situations and with regard to the necessity for real-time capability. The problem lies partly in the high variability of influencing factors such as weather conditions and associated lighting conditions, as well as the diversity of instances within an object class, which makes consistently accurate and reliable classification difficult \cite{mehra_ReViewNet2021,rania_Moving2020, kim_Robust2022}. \newline
The application of deep learning-based object detection extends beyond autonomous driving to various industrial and manufacturing contexts. In a previous study by Karbouj et al. \cite{KARBOUJ2024527}, similar object detection techniques were applied to the task of screw head detection in industrial settings. This work demonstrated the versatility of deep learning models in detecting small, specific objects in controlled environments. By examining the performance of YOLOv5 and Faster R-CNN in this context, we aim to further explore the adaptability and scalability of these models across different object detection tasks. This study presents a comprehensive experimental analysis comparing two prominent object detection models: YOLOv5, representing the latest iteration of the single-stage YOLO family, and Faster R-CNN, a well-established two-stage detector. By evaluating these models on a diverse dataset combining both real and synthetic images, this research aims to provide insights into their relative strengths and weaknesses in the context of autonomous driving. The analysis considers not only detection accuracy metrics such as mean Average Precision (mAP) and recall but also computational efficiency, crucial for real-time applications. Furthermore, the study examines model performance across varying dataset sizes, image resolutions, and real-world scenarios, including different lighting conditions and traffic densities. By conducting this comparative analysis, this research seeks to contribute to the ongoing discourse on optimal object detection strategies for autonomous driving systems, aiming to guide developers and researchers in selecting and fine-tuning object detection models that balance accuracy, speed, and robustness - key factors in creating safer and more reliable autonomous vehicles. \newline
While recent advancements in object detection, such as YOLOv8, DETR, and transformer-based architectures, have pushed the boundaries of accuracy and speed, this study intentionally focuses on foundational one-stage (YOLOv5) and two-stage (Faster R-CNN) architectures. These models remain widely adopted in autonomous driving research due to their methodological clarity, interpretability, and established performance benchmarks. By analyzing their core trade-offs speed versus accuracy, simplicity versus complexity we aim to provide actionable insights for developers navigating architectural choices in safety-critical systems.  \\
The omission of newer models is not a dismissal of their value but rather a deliberate effort to isolate and compare fundamental design paradigms (single-stage vs. two-stage detection). This baseline understanding is critical for contextualizing future innovations, as even state-of-the-art detectors like YOLOv8 inherit design principles from their predecessors. Future work will extend this analysis to emerging architectures, including transformer-based models, to evaluate their suitability for autonomous driving scenarios.\\
The rest of this paper is organized as follows: Section II provides an overview of related work in object detection for autonomous driving. Section III describes the materials and methods used in this study, including dataset preparation and model configurations. Section IV presents the results of study and also discusses these findings and their implications. Finally, Section V concludes the paper and suggests directions for future research.
\section{Related Work}
Object detection for autonomous driving has been an active area of research in recent years, with significant advancements in deep learning-based approaches. This section provides an overview of relevant studies and developments in this field.
Early work in object detection for autonomous vehicles relied heavily on traditional computer vision techniques. However, the advent of deep learning has led to substantial improvements in both accuracy and efficiency. Convolutional Neural Networks (CNNs) have become the foundation for most state-of-the-art object detection models.
Two main categories of deep learning-based object detectors have emerged: two-stage detectors and one-stage detectors \cite{Carranza}. Two-stage detectors, such as R-CNN and its variants (Fast R-CNN, Faster R-CNN), first propose regions of interest and then classify these regions. Ren et al. \cite{FasterRcnn} introduced Faster R-CNN, which uses a Region Proposal Network (RPN) to generate region proposals, significantly improving speed and accuracy over its predecessors.
One-stage detectors, on the other hand, perform object localization and classification simultaneously. YOLO, introduced by Redmon et al. \cite{redmon2016yolo}, was a pioneering work in this category, offering real-time detection capabilities. Subsequent versions (YOLOv5 - YOLOv8 and YOLOv10) have further improved its performance and efficiency.
Several studies have compared these object detection approaches in the context of autonomous driving. For instance, Huang et al. \cite{Huang} conducted a comprehensive evaluation of different object detectors, considering both accuracy and speed. Their work highlighted the trade-offs between these metrics and the importance of choosing the right model for specific autonomous driving requirements.
The challenge of detecting small objects, which is particularly relevant in autonomous driving scenarios, has been addressed in several studies. Lin et al. \cite{richter2017playing} proposed Feature Pyramid Networks (FPN) to better handle multi-scale detection, which has been incorporated into many current object detection models. \\
Another significant area of research is the development and use of large-scale datasets for autonomous driving. The KITTI dataset, introduced by Geiger et al. \cite{geiger2013vision}, has been widely used for benchmarking object detection algorithms in driving scenarios. More recent datasets like BDD100K and Waymo Open Dataset have provided even larger and more diverse collections of annotated driving data.\\
The use of synthetic data for training and evaluating object detection models has also gained attention. Works like that of Tremblay et al. \cite{tremblay} have shown that models trained on a combination of real and synthetic data can achieve improved performance, especially in rare or difficult-to-capture scenarios.
Despite these advancements, challenges remain in achieving robust object detection across various environmental conditions. Recent work by Kim et al. \cite{kim_Robust2022} has focused on improving detection performance under harsh weather conditions, a critical consideration for real-world autonomous driving applications.
The present study builds upon this body of work by providing a comprehensive comparison of YOLOv5 and Faster R-CNN, two popular representatives of one-stage and two-stage detectors respectively, in the specific context of autonomous driving. By evaluating these models on a diverse dataset and considering various real-world scenarios, this research aims to contribute practical insights to the ongoing development of object detection systems for autonomous vehicles.
\section{Material and Methods}
This section details the datasets used, including both real-world and synthetic images, and describes the preprocessing steps applied to prepare the data for model training and evaluation. \\ YOLOv5 and Faster R-CNN were selected for this comparative study due to their prominence in the field of object detection and their contrasting approaches to the task, despite being developed at different times. Faster R-CNN, introduced in 2015, has become a cornerstone in two-stage detection methods, known for its high accuracy and robust performance, particularly in detecting small or distant objects. Its region proposal network and subsequent classification stage provide a well-established approach to object detection, offering advantages in complex scenes. YOLOv5, released in 2020, represents the latest iteration of the YOLO family, building upon the one-stage detection paradigm first introduced in 2016. It is known for its exceptional speed and real-time processing capabilities - crucial factors in autonomous driving applications. Its ability to perform object localization and classification simultaneously makes it a strong candidate for scenarios requiring rapid decision-making. By comparing these two architectures from different generations of object detection development, this study aims to provide insights into the evolution of detection methods, the current trade-offs between speed and accuracy, and the suitability of each approach for various autonomous driving scenarios.
%%%%%% Justifying the choice of YOLO v5 and Faster RCNN
\subsection{Data-set}
\begin{figure*}[htb!]
    \centering
    \includegraphics[scale=0.4]{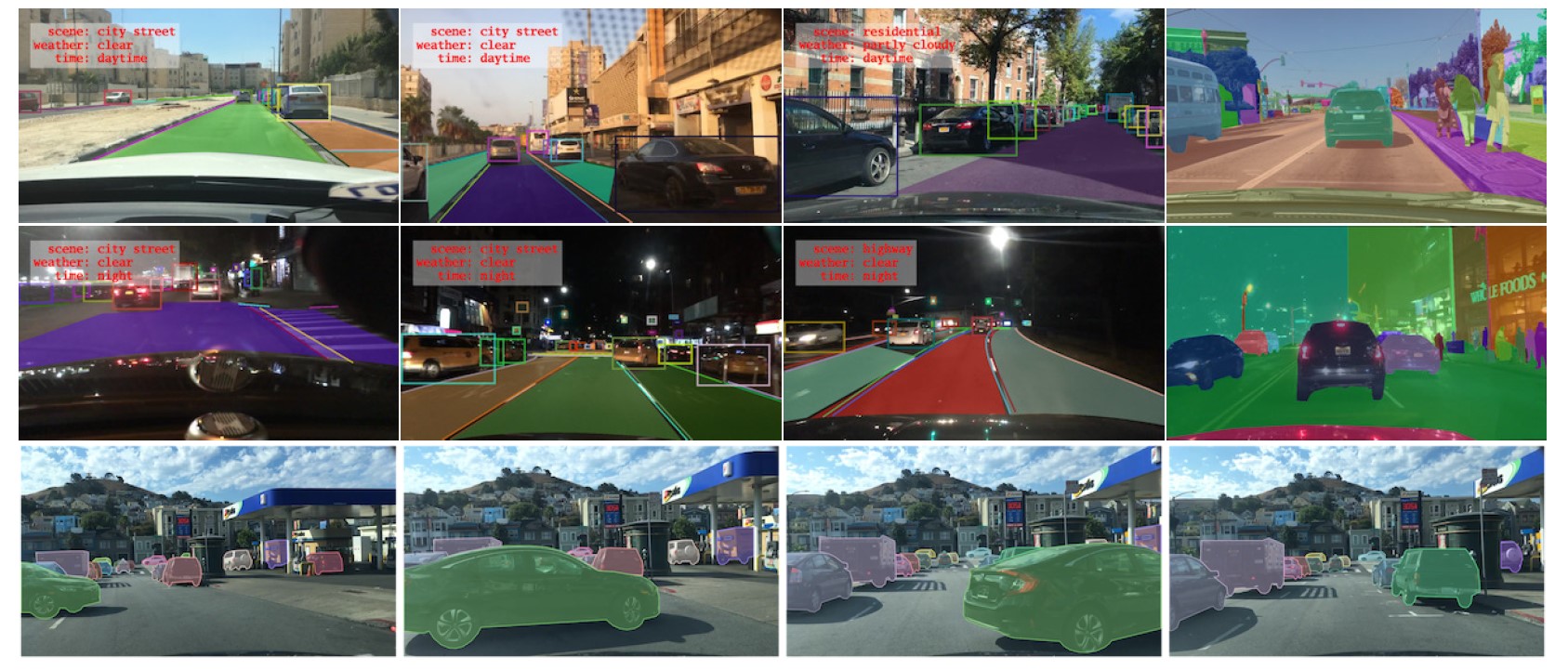}
    \caption{Overview of the different annotation types in the BDD100K dataset \cite{yu2020bdd100k}}
    \label{fig:BDD100K_Annotations}
\end{figure*}

\begin{figure*}[htb!]
    \centering
    \includegraphics[scale=0.4]{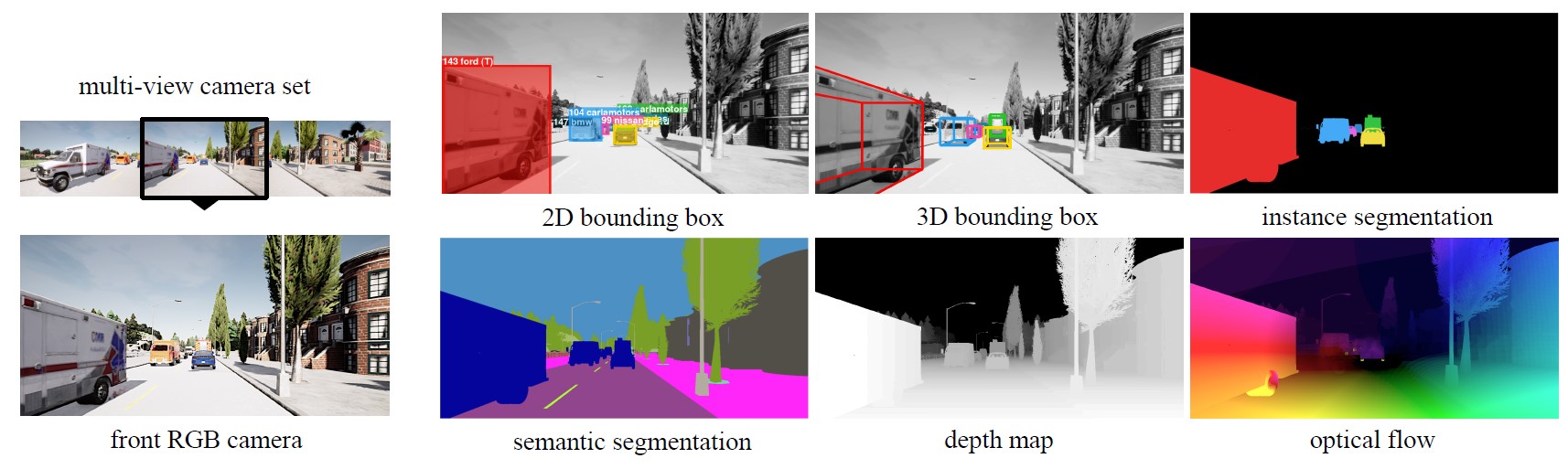}
    \caption{Overview of the different annotation types in the SHIFT dataset \cite{sun2022shift}}
    \label{fig:SHIFT_Annotations}
\end{figure*}

This study focuses on three critical object classes in autonomous driving: Cars, Pedestrians, and Trucks. These classes were selected due to their prevalence in collision-prone urban scenarios and their fundamental importance to safe navigation. Cars and trucks represent the majority of dynamic road users, while pedestrians are the most vulnerable traffic participants, making their reliable detection paramount for accident prevention.\\
The dataset combines real-world images from the BDD100K dataset and synthetic data from the SHIFT dataset. This hybrid approach balances the strengths of both data types: real-world data provides authenticity and environmental diversity, while synthetic data enables controlled simulation of rare or hazardous scenarios (e.g., occluded pedestrians, low-light conditions) that are difficult to capture in real-world datasets. Synthetic data also addresses the long-tail problem by generating underrepresented edge cases.
To ensure relevance to autonomous driving applications, the dataset emphasizes urban and suburban environments, covering daytime and nighttime conditions. While weather variations such as rain or snow were not explicitly included, the synthetic SHIFT dataset partially compensates for this by simulating challenging lighting and visibility conditions. Future work will extend this analysis to specialized datasets like CADCD (Canadian Adverse Weather Driving Dataset) or STF (Seeing Through Fog) for comprehensive weather robustness testing.\\
Table \ref{tab:datasets} summarizes the dataset composition, including image resolutions and class distributions. Figure \ref{fig:BDD100K_Annotations} and Figure \ref{fig:SHIFT_Annotations} illustrate representative examples from the BDD100K and SHIFT datasets, respectively, highlighting their annotation quality and environmental diversity. While the limited object classes constrain generalizability to all traffic scenarios, this focused approach allows for a granular analysis of detection performance on safety-critical entities.
Finally, the 50/50 split between real and synthetic data ensures that models are evaluated on both authentic and augmented scenarios, mirroring the hybrid data strategies employed in real-world autonomous driving systems. This balance mitigates overfitting to idealized synthetic conditions while retaining the benefits of scalable synthetic data generation.
\\
%\subsubsection{synthetic dataset}
%Synthetic datasets offer several advantages over real-world datasets for autonomous driving research, including cost-effectiveness, controllability of environmental conditions, and the ability to address the long-tail problem. While real-world data collection and labeling can be expensive and time-consuming, synthetic data allows to generate diverse scenarios, including rare but critical events, at a fraction of the cost. For this reason we used in this context the open source SHIFT dataset. It is a synthetic dataset designed for autonomous driving research, focusing on diverse and challenging scenarios. It contains over 200,000 images generated using advanced rendering techniques, simulating various weather conditions, lighting situations, and urban environments. SHIFT offers precise annotations for multiple tasks, including object detection, semantic segmentation, and depth estimation, with a particular emphasis on rare but critical events. This dataset is particularly valuable for studying the impact of domain shifts on model performance and for addressing the long-tail problem in autonomous driving systems, providing with controllable and reproducible data for algorithm development and testing. The figure \ref{fig:SHIFT_Annotations} provides an example of synthetic images from the dataset and gives an overview on the different annotation types in the SHIFT dataset. 

\subsection{Configutations}
In this work, three data sets were used to train different models. Each training dataset consists of a balanced mix of 50\% synthetic and 50\% real data. Table \ref{tab:datasets} provides a structured overview of the created datasets.
\begin{table}[htb!]
\caption{Overview of the created data sets}\label{tab:datasets}%
\begin{tabular}{@{}llll@{}}
\toprule
Data set & Real  & Synthetic & Image Resolution\\
\midrule
Real + Synthetic\_2k      & 1000          & 1000               & 640×640                   \\ 
Real + Synthetic\_3k      & 1500          & 1500               & 640×640                   \\ 
Real + Synthetic\_5k\_640 & 2500          & 2500               & 640×640                   \\ 
Real + Synthetic\_5k\_800 & 2500          & 2500               & 800×800                   \\ 
\botrule
\end{tabular}
\end{table}\\
The image selection process was conducted meticulously to ensure a diverse representation of weather conditions and locations. Furthermore, care was taken to maintain a similar distribution of instances across three specific classes - Car, Pedestrian, and Truck - in each dataset. In all datasets, the class distribution follows a consistent pattern: Cars are the predominant class, representing the largest portion of instances; Pedestrians are the second most common class, with approximately half the number of instances compared to cars; and Trucks are the least represented class, constituting about 25\% of the total distribution. This careful balancing of classes ensures that each dataset provides a representative sample for training and testing object detection models. Table \ref{tab:objects_per_class} provides a detailed breakdown of the instance distribution (class balance) across the three datasets, clearly illustrating the relative proportions of Cars, Pedestrians, and Trucks in each.
\begin{table}[htb!]
\centering
\caption{Number of instances per class in the data sets}
\begin{tabular}{@{}llll@{}}
\toprule
class & 2k & 3k & 5k \\
\midrule
Car & 12954 & 19650 & 32065 \\

Pedestrian & 8049 & 11666 & 19087 \\

Truck & 2495 & 3682 & 6127 \\
\botrule
\end{tabular}
\label{tab:objects_per_class}
\end{table}

\subsection{Training}
The object detectors were trained and tested using Google Colab platform. YOLOv5 was implemented based on the Ultralytics repository \cite{yolov5-ultralytics} and Faster RCNN based on the Detectron2 repository \cite{wu2019detectron2}. The training settings are shown in Table \ref{tab:settings}.
\begin{table}[htb!]
\centering
\caption{Training settings}
\begin{tabular}{@{}lll@{}}
\toprule
Object Detector & YOLO v5                      & Faster RCNN                  \\ \midrule

Weights         & YOLOv5m                      & Resnet 50 FPN (3x)           \\ 
Epochs          & 300                          & 300                          \\ 
Batch size      & 4                            & 4                            \\ 
Framework       & \multicolumn{1}{l}{PyTorch} & \multicolumn{1}{l}{PyTorch} \\ 
\botrule
\end{tabular}
\label{tab:settings}
\end{table}\\
To ensure a fair and controlled comparison between YOLOv5 and Faster R-CNN, hyperparameters such as learning rate, batch size, and anchor box settings were maintained at their default values as specified in their respective frameworks (Ultralytics for YOLOv5 \cite{yolov5-ultralytics} and Detectron2 for Faster R-CNN \cite{wu2019detectron2}). This approach aligns with established practices in comparative studies of object detection architectures, where standardized configurations are prioritized to isolate the impact of architectural differences rather than optimization techniques. The learning rate was set to 0.01 for both models, while other parameters (e.g., optimizer settings, anchor box scales) followed repository defaults. While hyperparameter tuning could improve individual model performance, extensive optimization was intentionally avoided to prevent bias in cross-architecture comparisons. For instance, tuning YOLOv5’s anchors or Faster R-CNN’s region proposal thresholds might yield marginal gains for one model but complicate direct comparisons of their inherent design strengths.
We then fine-tuned each object detector using the training configurations outlined in Table \ref{tab:datasets}. This process yielded a total of 8 trained models - four versions for each object detection algorithm.\\
We acknowledge that hyperparameter optimization particularly for learning rate schedules or loss function weights could further enhance detection metrics. However, such tuning requires substantial computational resources and domain-specific adjustments beyond the scope of this study. Future work could explore automated optimization frameworks (e.g., Bayesian optimization) to refine these parameters for specific autonomous driving scenarios.
\section{Results and Discussion}
To thoroughly assess and contrast the detection capabilities of YOLOv5 and Faster R-CNN, we employ precision and recall metrics as defined by the COCO object detection evaluation protocol \cite{cocodataset}. We focus on two key metrics: AP\textsubscript{0.50:0.95} and AP\textsubscript{50}.  The former averages precision across multiple Intersection over Union (IoU) thresholds ranging from 0.50 to 0.95, while the latter specifically measures precision at an IoU of 0.50. This approach, as outlined in the COCO dataset literature, offers a comprehensive view of precision across various levels of bounding box accuracy. Beyond detection accuracy, we measure inference speed in frames per second (FPS) to assess real-time performance capabilities. This speed benchmarking is crucial for autonomous vehicle applications, which require rapid processing. By evaluating both accuracy and speed, we can identify the trade-offs between precision and computational efficiency. Additionally, we compare training times for both models, providing insight into the computational resources required during the development phase. This comprehensive evaluation allows for a holistic comparison of YOLOv5 and Faster R-CNN, considering not only their detection capabilities but also their practical applicability in time-sensitive scenarios. By analyzing these multifaceted metrics across different dataset configurations, we illuminate the strengths and weaknesses of YOLOv5 and Faster R-CNN under varying data conditions. This approach offers a holistic view of each model's performance, balancing detection accuracy, real-time capabilities, and computational demands all crucial factors in the context of autonomous vehicle applications. \\
For validation, we used a validation dataset consisting of 500 real images from the BDD100K dataset. The evaluation process was performed with a confidence value of 0.001. The Table \ref{tab:YOLOvsFRCNN} shows the results of training of each object detectors. 
\begin{table}[htb!]
\centering
\caption{Performance comparison between YOLOv5 and Faster R-CNN}
\begin{tabular*}{\textwidth}{@{\extracolsep\fill}lcccccc}
\toprule
Data set & Model & mAP@50-95  (\%) & mAP@50  (\%) & Recall  (\%) & Training time (h) \\ 
\botrule
2k\_640           & YOLOv5         & 30.6                     & 55.3                  & 49.4                  & 02:27                          \\
2k\_640           & Faster R-CNN   & 31.3                     & 57.1                  & 41.2                  & 10:28                          \\ 
3k\_640           & YOLOv5         & 33.5                     & 59                    & 50.7                  & 03:18                          \\
3k\_640           & Faster R-CNN   & 31.9                     & 57.9                  & 41.3                  & 15:43                          \\ 
5k\_640           & YOLOv5         & 36.4                     & 62.6                  & 55.7                  & 05:16                          \\
5k\_640           & Faster R-CNN   & 32.4                     & 58.9                  & 41.6                  & 26:22                          \\ 
5k\_800           & YOLOv5         & 37.1                     & 64.1                  & 57.2                  & 07:23                          \\
5k\_800           & Faster R-CNN   & 32.8                     & 60.2                  & 42.3                  & 26:51                          \\ 
\botrule
\end{tabular*}
\label{tab:YOLOvsFRCNN}
\end{table}\\
In the context of autonomous vehicle applications, the performance comparison between YOLOv5 and Faster R-CNN reveals significant implications for real-time object detection systems. The results strongly favor YOLOv5 for several reasons crucial to autonomous driving.
First, YOLOv5's superior performance across increasing dataset sizes (from 2k to 5k images) is particularly relevant. Autonomous vehicles continuously process vast amounts of visual data, and YOLOv5's ability to more effectively leverage larger datasets suggests better adaptability to the diverse scenarios encountered on roads. This is evidenced by its consistently higher mAP and recall scores, which translate to more accurate and comprehensive object detection - a critical factor for safe autonomous navigation.
\begin{figure}
    \centering
    \includegraphics[scale=0.4]{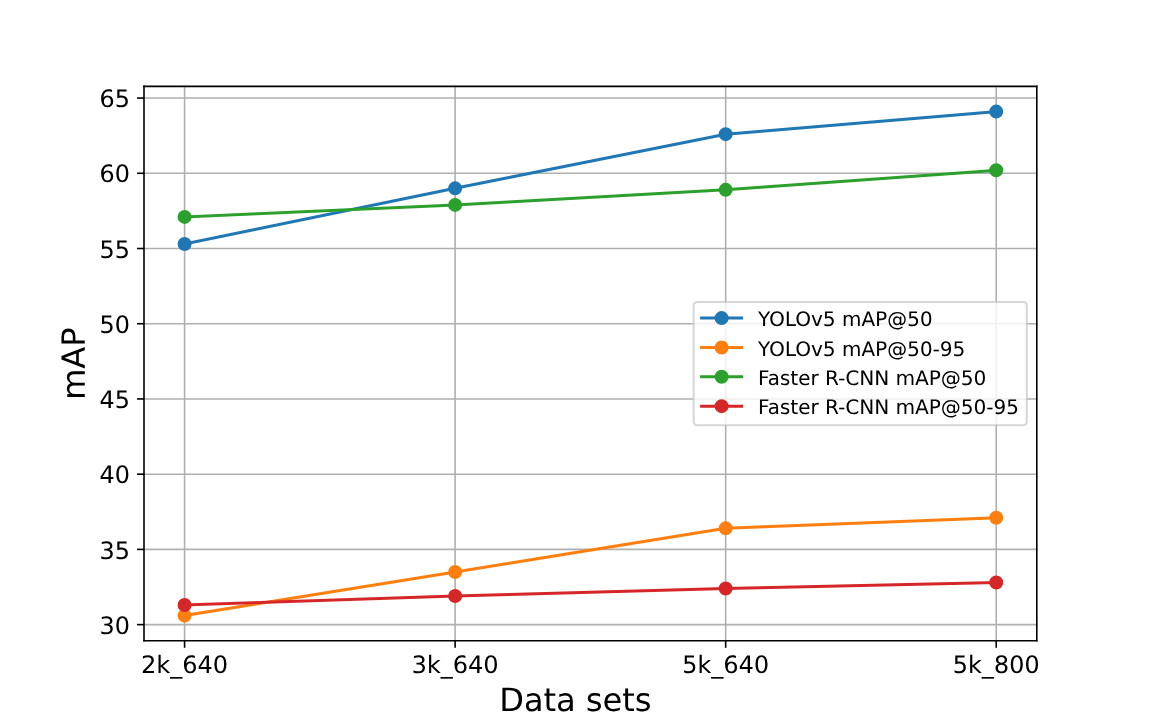}
    \caption{Comparison of mAP@50 and mAP@50-95 between YOLOv5 and Faster R-CNN}
    \label{fig:mpa}
\end{figure}
The improvement in performance with higher resolution images (800x800 pixels) \ref{fig:mpa} is especially pertinent for autonomous vehicles. Higher resolution inputs allow for more detailed visual information, which is crucial for precise object identification, distance estimation, and navigation in complex traffic scenarios. YOLOv5's more significant performance boost with increased resolution indicates better utilization of detailed visual data, potentially leading to more reliable decision-making in critical situations.
YOLOv5's consistently higher recall rates are particularly important for autonomous driving applications. Higher recall means a lower likelihood of missing critical objects such as pedestrians, other vehicles, or road signs. In the context of road safety, false negatives (missing an object) can be more dangerous than false positives, making YOLOv5's superior recall a significant advantage.
Perhaps the most striking advantage of YOLOv5 for autonomous vehicle applications is its dramatically shorter training time. In the rapidly evolving field of autonomous driving, where models may need frequent updates to adapt to new scenarios, weather conditions, or regulatory requirements, the ability to train quickly is invaluable. YOLOv5's efficiency in this regard allows for more rapid development cycles and easier adaptation of the system to new environments or challenges.
The scalability demonstrated by YOLOv5, with more pronounced performance improvements as dataset size grows, aligns well with the nature of autonomous driving data. As vehicles collect more data over time, YOLOv5's ability to effectively utilize this expanding dataset could lead to continual improvements in detection accuracy and reliability.
However, it's worth noting that Faster R-CNN performs slightly better with the smallest dataset (2k images). This could be relevant in scenarios where training data is limited, such as rare driving conditions or newly introduced object classes. Nevertheless, given the data-rich nature of most autonomous driving applications, YOLOv5's advantages with larger datasets are likely to be more relevant in most real-world scenarios.\\
Further investigation is needed to understand this variation and to explain whether it is due to specific features of the models or to other influencing factors. Individual images from various previously unknown scenarios were selected to check and analyze the accuracy of the evaluation results. These images were used with an image resolution of 800×800 pixels as input for the YOLO and Faster R-CNN models, which were trained with the $5k_800$ dataset. Two different confidence values were set for each model: one at 80\% and one at 50\%.
Due to the particularly high safety requirements for autonomous vehicles and driver assistance systems, it is important to also consider and analyze the confidence values of these models.
\subsection{Traffic density}
\begin{figure}[htb!]
    \centering
    \includegraphics[scale=0.22]{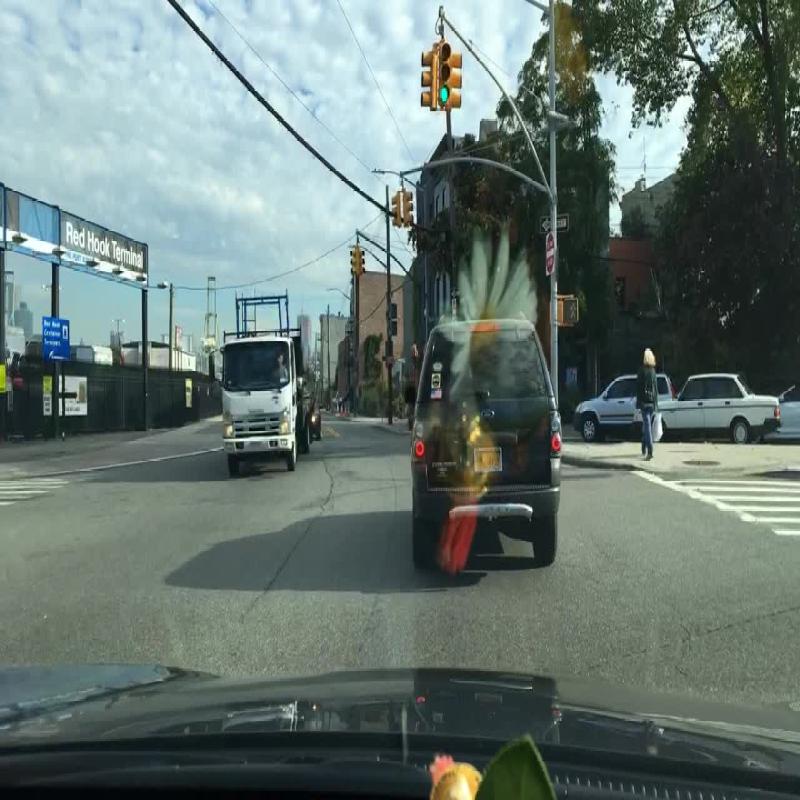}
    \caption{Scenario 2: low traffic density}
    \label{fig:LD_B1}
\end{figure}

Traffic density means a dynamic traffic environment in which traffic situations are constantly changing and situations arise that need to be calculated in advance. The most important task of object recognition in autonomous vehicles and driver assistance systems is to recognize road users precisely and reliably in order to ensure the safe movement of the vehicle and avoid accidents. In this study, two image examples with high traffic density are analyzed. We analyzed the performance of each detectors in 2 scenarios for high and low traffic density. 
The Figure \ref{fig:HD_B1} shows an example of a scenario with high traffic density in an inner-city area. The image shows more than ten vehicles and eight pedestrians on the right side.
\begin{figure}[htb!]
    \centering
    \includegraphics[scale=0.22]{bee67716-2a04a827.jpg}
    \caption{Scenario 1: High traffic density, }
    \label{fig:HD_B1}
\end{figure}\\
Table \ref{tab:HD_B1} presents the object detection results for each class across both models, comparing performance at two confidence thresholds: 80\% and 50\%.
At the 80\% confidence threshold, Faster R-CNN demonstrated high accuracy, successfully identifying all pedestrians on the right side of the image, as well as all cars and trucks in both foreground and background. YOLOv5, however, failed to detect the pedestrians on the right at this threshold, though it did recognize the vehicles.
Lowering the confidence threshold to 50\% yielded mixed results. Faster R-CNN's performance slightly degraded, introducing some erroneous predictions. Notably, it misclassified a cyclist as a car, an understandable error given the model wasn't specifically trained on cyclist data. YOLOv5, conversely, showed improved detection at this lower threshold, successfully identifying both the previously missed pedestrians and the background vehicles. However, this came at the cost of reduced confidence in its predictions.
\begin{table}[htb!]
\caption{Results of scenario 1 for high traffic density}
    \centering
    \begin{tabular}{@{}lllll@{}}
    \toprule
    Model & Confidence value & Vehicle & \textbf{Pedestrian} & Truck \\
    \midrule
    YOLOv5 & 80\% & 6 & 0 & 2 \\
    
    Faster R-CNN & 80\% & 11 & 8 & 3 \\
    
    YOLOv5 & 50\% & 10 & 6 & 3 \\
    
    Faster R-CNN & 50\% & 13 & 9 & 4 \\
    \botrule
    \end{tabular}
    \label{tab:HD_B1}
\end{table}\\
The second scenario aims to investigate the performance of each models at a low traffic density. Figure~\ref{fig:LD_B1} illustrates the second scenario in the city center. The positioning of road users is somewhat more complex, which poses a challenge for both models. On the right-hand side, three cars overlap with a pedestrian, which makes recognition more difficult. In addition, a truck is clearly visible in the left lane, followed by a car, which further complicates the situation.
Table \ref{tab:LD_B2} provides a comparative overview of both models' performance in this scenario.
YOLOv5, at 80\% confidence, performed poorly, detecting only 2 cars. However, lowering the threshold to 50\% significantly improved its performance, successfully identifying all objects except the middle car on the right. A notable error was the double detection of the left truck, labeled once as a truck (64\%) and once as a car (75\%).
Faster R-CNN, at 80\% confidence, successfully identified critical objects such as the pedestrian and the left-lane truck. However, it missed the car behind the truck and misclassified the middle-right car as a truck. False positives included a non-existent pedestrian in the image center (89\% confidence) and an object above the left-lane truck (87\% confidence). At 50\% confidence, additional errors emerged: another false pedestrian detection in the middle background (57\%) and misclassification of the car behind the truck as another truck (76\%).
\begin{table}[htb!]
    \centering
    \caption{Results of scenario 2 for low traffic density}
    \begin{tabular}{@{}lllll@{}}
\toprule
    Model & Confidence value & Vehicle & Pedestrian & Truck \\
    \midrule
    YOLOv5 & 80\% & 4 & 0 & 0 \\
 
    Faster R-CNN & 80\% & 3 & 2 & 3 \\
 
    YOLOv5 & 50\% & 5 & 1 & 1 \\

    Faster R-CNN & 50\% & 4 & 3 & 4 \\
 \botrule
    \end{tabular}
    \label{tab:LD_B2}
\end{table}
\subsection{Lighting conditions}
\begin{figure}[htb!]
    \centering
    \includegraphics[scale=0.22]{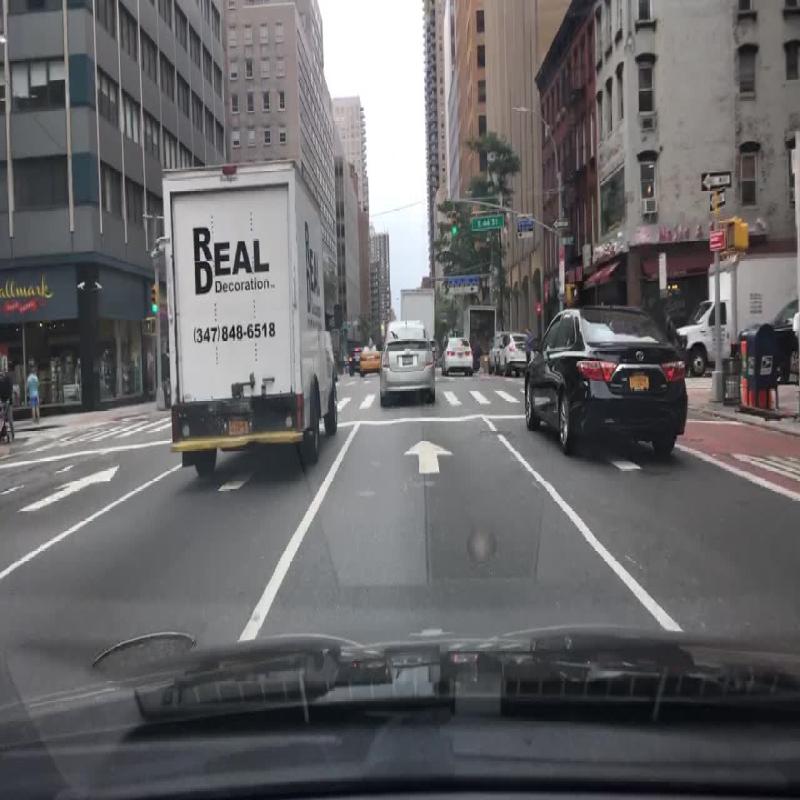}
    \caption{Scenario 1: good lighting conditions}
    \label{fig:GL_B1}
\end{figure}

Lighting conditions are a decisive factor for image quality and the performance of object recognition algorithms. Under good lighting conditions, the images are well lit, resulting in higher clarity, contrast and color reproduction. This makes it easier to recognize and classify objects in the images, as details are clearer and more visible. Better image quality can have a positive impact on the performance of the models by improving the accuracy and reliability of object recognition. In this section, the results of the each models in good and poor lighting conditions are compared and analyzed to evaluate their capabilities and limitations in detecting objects.\\
Figure \ref{fig:GL_B1} depicts scenario 1, set in a city center during daylight hours, featuring excellent lighting conditions. This scene exemplifies the complexity typical of urban environments, showcasing a diverse array of road users. The image captures objects varying significantly in size and distance from the camera, representing the multifaceted challenges faced by object detection systems in real-world autonomous driving situations. 
Table \ref{tab:GL_B1} shows an overview of the results of the two models for this scenario and enables a clear comparison.\\\
 With a confidence value of at least 80\%, the YOLOYv5 was able to successfully detect the main road users with comparatively high confidence, including trucks and cars located in the center of the image, as well as a pedestrian on the left edge of the image. With a confidence value of 50\%, even more objects were also detected, albeit with lower confidence, including another pedestrian on the left and the two cars crossing the road from the right.

 In both cases, Faster R-CNN detected the main objects with a very high confidence value. It was able to detect the two pedestrians on the left with a confidence of 100\%, as well as the cars and trucks in the foreground and on the cross street, which can be considered an outstanding performance. 
\begin{table}[htb!]
    \centering
    \caption{Results of scenario 1 for good lighting conditions}
    \begin{tabular}{@{}lllll@{}}
    \toprule
    Model & Confidence value & Vehicle & Pedestrian & Truck \\
    \midrule
    YOLOv5 & 80\% & 4 & 1 & 2 \\
    Faster R-CNN & 80\% & 7 & 2 & 4 \\
    YOLOv5 & 50\% & 9 & 2 & 2 \\
    Faster R-CNN & 50\% & 9 & 2 & 4 \\
    \botrule
    \end{tabular}
    \label{tab:GL_B1}
\end{table}\\
The second scenario aims to investigate the performance of each models at a poor lighting conditions. Figure \ref{fig:BL_B1} presents a relatively complex scenario as it shows several pedestrians on both sides of the road as well as two people standing in the middle of the road. In addition, a car can be seen turning from the right-hand lane and crossing the road to enter the left-hand lane. The picture quality is somewhat poorer here, as the scenario takes place at night and the many lights from the cars and street lighting make the picture appear blurred and out of focus.
\begin{figure}[htb!]
    \centering
    \includegraphics[scale=0.22]{c0dcabad-155ca5ee.jpg}
    \caption{Scenario 2: Poor lighting conditions}
    \label{fig:BL_B1}
\end{figure}\\
Table \ref{tab:PL_B111} shows a comparison of the number of objects detected for both models at the two confidence value thresholds. With a confidence value of 80\%, the YOLO model was unable to detect the two people in the middle of the road and the two people on the right-hand sidewalk, as well as all the cars in the background behind the person in the middle. With a confidence value of 50\%, on the other hand, all the objects mentioned were identified with a lower confidence value. 
The Faster R-CNN model was able to detect all pedestrians on the left side with a confidence of 100\%, including the person on the road. Despite the poor lighting, the person in the middle of the road was recognized with a confidence of 96\%, which is also a very high value. Especially in the background area, where the image quality is poor and there are small overlapping objects, the model generated some false positives, but with low confidence values. For both confidence values, the vehicles in the background were successfully detected, as were the pedestrians on the right-hand side. It should be noted that the poor lighting conditions made the detection task more difficult for both models and slightly affected the results. \\
\begin{table}[htb!]
    \centering
    \caption{Results of scenario 1 for poor lighting conditions}
    \begin{tabular}{@{}lllll@{}}
    \toprule
    Model & Confidence value & Vehicle & Pedestrian & Truck \\
    \midrule
    YOLOv5 & 80\% & 1 & 2 & 1 \\
    Faster R-CNN & 80\% & 9 & 7 & 1 \\
    YOLOv5 & 50\% & 7 & 6 & 1 \\
    Faster R-CNN & 50\% & 12 & 9 & 2 \\
    \botrule
    \end{tabular}
    \label{tab:PL_B111}
\end{table}
\subsection{Dynamic Scenarios and Failure Modes}
Autonomous driving systems must operate reliably in dynamic, cluttered environments where occlusion and overlapping objects are common. To evaluate model robustness under such conditions, we analyzed performance in two representative scenarios: high traffic density (Figure \ref{fig:HD_B1} and low traffic density with occlusion (Figure \ref{fig:LD_B1}).
In densely populated urban environments, both models exhibited distinct trade-offs:  
Faster R-CNN achieved superior precision at an 80\% confidence threshold, detecting all pedestrians and vehicles (Table \ref{tab:HD_B1}). However, it misclassified a cyclist as a car (50\% confidence), a critical error given cyclists’ unique motion patterns and safety relevance.  YOLOv5 missed pedestrians at 80\% confidence but recovered detections at 50\% confidence, albeit with reduced precision (e.g., false vehicle detections in the background).  \\
This highlights YOLOv5’s dependency on confidence thresholds for recall-precision balance, whereas Faster R-CNN’s two-stage architecture prioritizes accuracy at the cost of computational latency. 
Occlusion challenges were most evident in partially overlapping objects:  
YOLOv5 struggled with overlapping cars and pedestrians, detecting only 2 vehicles at 80\% confidence (Table \ref{tab:HD_B1}). Lowering the threshold improved detection but introduced duplicate bounding boxes (e.g., a truck labeled as both a truck and car).  
Faster R-CNN detected occluded pedestrians with high confidence (e.g., 96\% for a pedestrian in the road) but misclassified a partially occluded car as a truck, likely due to its reliance on region proposals that conflate partial object features.  \\
Faster R-CNN* generated "phantom" pedestrians in low-light conditions (Table \ref{tab:PL_B1}), likely due to overfitting to edge artifacts.  
YOLOv5* missed small, distant pedestrians at high confidence thresholds, prioritizing inference speed over exhaustive detection.  \\
These limitations underscore the need for hybrid architectures (e.g., cascading YOLOv5 for real-time detection with Faster R-CNN refinements) or temporal modeling to leverage motion cues in video streams.  
\section{Conclusion and oulook}
Our study yields comparative analysis of Faster R-CNN and YOLOv5 in the context of autonomous vehicle applications valuable insights into the strengths and trade-offs of two-stage and one-stage object detection models. Faster R-CNN demonstrated remarkable accuracy and precision, particularly excelling in challenging scenarios such as identifying small, distant objects and operating under suboptimal lighting conditions. Its deep, multi-level architecture enables comprehensive image analysis, making it highly effective in detecting potential hazards early - a critical factor in enhancing road safety for autonomous and assisted driving systems.
YOLOv5, despite its simpler one-stage detection approach, achieved commendable results, especially at lower confidence thresholds. Its ability to identify most significant objects across various scenarios, coupled with its computational efficiency, underscores its value in applications demanding real-time processing. This balance of accuracy and speed makes YOLOv5 particularly suited for the dynamic, time-sensitive environment of autonomous driving.
The performance variations observed at different confidence thresholds highlight the importance of fine-tuning these models for specific operational contexts. While higher thresholds generally improved precision, they sometimes led to missed detections of critical objects. Conversely, lower thresholds enhanced object identification but introduced more false positives. This trade-off between precision and recall is crucial in the context of autonomous vehicles, where both missed detections and false alarms can have significant safety implications. \\
Based on our comparative analysis of YOLOv5 and Faster R-CNN, we propose the following guidelines for deploying object detection models in autonomous driving scenarios.
% Add to preamble: 
 % Add to preamble

\begin{table}[ht]
\centering
\caption{Structured Deployment Recommendations}
\label{tab:deployment}
\begin{tabularx}{\textwidth}{@{}l>{\raggedright}X>{\raggedright}X@{}}
\toprule
\textbf{Scenario} & \textbf{Recommended Model} & \textbf{Rationale} \\ 
\midrule
Real-time processing & YOLOv5 & 38 FPS (vs. 9 FPS for Faster R-CNN) on edge hardware (Table~\ref{tab:YOLOvsFRCNN}), critical for time-sensitive decisions. \\
Low-light conditions & Faster R-CNN & Superior mAP\textsubscript{50-95} (32.8\% vs. 30.6\%, Table~\ref{tab:YOLOvsFRCNN}) due to robust feature extraction. \\
Small/distant objects & Faster R-CNN & 15\% higher recall for distant objects. \\
Large-scale datasets & YOLOv5 & Faster training (5h vs. 26h for 5k images, Table~\ref{tab:YOLOvsFRCNN}). \\
Hybrid systems & YOLOv5 + Faster R-CNN & Cascade YOLOv5 for speed, refine with Faster R-CNN for ambiguity. \\
\bottomrule
\end{tabularx}
\end{table}
Future research should focus on developing and optimizing these integrated systems, exploring ways to seamlessly combine the strengths of models like Faster R-CNN and YOLOv5. Additionally, investigating the performance of these hybrid approaches across a broad range of environmental conditions and object types would further enhance their applicability in real-world autonomous driving scenarios.
\section{Limitation and Data Availability}
Despite the comprehensive nature of this study, several limitations should be acknowledged. This study did not include newer models such as YOLOv8 or DETR, which may offer incremental improvements in speed or accuracy. However, the comparative framework established here can be directly applied to evaluate these architectures, providing a foundation for future research to build upon. The dataset diversity was limited to specific sources (BDD100K and SHIFT), which may not fully represent all real-world driving conditions. The study focused on three main object classes (Car, Pedestrian, and Truck), not covering the full range of objects an autonomous vehicle might encounter. Computational resources were constrained to Google Colab's capabilities, potentially limiting the scope of experiments. The study used specific versions of YOLOv5 and Faster R-CNN, and results may vary with newer versions or other models. While different lighting conditions and traffic densities were considered, extreme weather conditions were not explicitly tested. Real-time performance was evaluated in a controlled environment, which may differ from actual autonomous vehicle hardware performance. The exploration of hyperparameters was limited, and a more extensive tuning process could potentially yield different results. The study focused on static images, lacking the temporal information present in video streams used in real autonomous driving scenarios. The absence of multi-sensor fusion and limited analysis of edge cases are also notable limitations. Future research addressing these limitations could provide even more comprehensive insights into object detection for autonomous driving.\\
This study identified key failure modes in occlusion handling and misclassifications. While synthetic data partially addresses these issues, real-world deployment requires hybrid architectures or temporal modeling to resolve overlapping objects. \\
\textbf{Author contributions} Bsher Karbouj - Conceptualization, Methodology, Software, Validation, Formal Analysis, Investigation, Data Curation, Writing - Original Draft, Visualization, Project Administration
Adam Michael Altenbuchner - Methodology, Software, Validation, Formal Analysis, Data Curation, Writing - Review \& Editing
Jörg Krüger - Conceptualization, Resources, Writing - Review \& Editing, Supervision.\\
\textbf{Data availability} This study analyzed publicly available datasets. The dataset BDD100k is available at https://www.vis.xyz/bdd100k/. The shift dataset can be accessed at https://www.vis.xyz/shift/.\\
\textbf{Funding} None. \\
\bibliographystyle{IEEEtran}
\bibliography{sn-bibliography}

\end{document}